# SilNet : Single- and Multi-View Reconstruction by Learning from Silhouettes

Olivia Wiles
ow@robots.ox.ac.uk
Andrew Zisserman
az@robots.ox.ac.uk

Visual Geometry Group
Department of Engineering Science
University of Oxford
Oxford, UK

**Abstract**

The objective of this paper is 3D shape understanding from single and multiple images. To this end, we introduce a new deep-learning architecture and loss function, SilNet, that can handle multiple views in an order-agnostic manner. The architecture is fully convolutional, and for training we use a proxy task of silhouette prediction, rather than directly learning a mapping from 2D images to 3D shape as has been the target in most recent work.

We demonstrate that with the SilNet architecture there is generalisation over the number of views – for example, SilNet trained on 2 views can be used with 3 or 4 views at test-time; and performance improves with more views.

We introduce two new synthetics datasets: a blobby object dataset useful for pre-training, and a challenging and realistic sculpture dataset; and demonstrate on these datasets that SilNet has indeed learnt 3D shape. Finally, we show that SilNet exceeds the state of the art on the ShapeNet benchmark dataset [6] at generating silhouettes in new viewpoints, and we use SilNet to generate novel views of the sculpture dataset.

## 1 Introduction

Inferring 3D shape from an image is one of the core problems of computer vision. Another of the many benefits of deep learning has been a resurgence of interest in this task. Many recent works have developed the idea of inferring 3D shape given a set of classes (e.g. cars, chairs, rooms) and a large dataset of synthetic 3D models of those classes for training [7, 10, 13, 29, 32, 37, 39, 40]. This modern treatment of class based reconstruction follows on from the pre-deep learning classic work of Blanz and Vetter for faces [2], and later for other classes such as semantic categories [19] or cuboidal room structures [11, 17].

In this paper we extend this area in two directions: first, we consider 3D shape inference from *multiple* images, rather than only a single one (though we consider this as well); second, we consider the quite generic class of 3D undulations – smooth curved surfaces – and apply this to the case of piecewise smooth textured sculptures. An example is shown in figure 1.

To achieve these extensions we introduce a deep learning architecture, *SilNet*, that can learn to encode 3D shape from one or more input images. The encoding can be used to generate new views or a 3D rendering by modifying only the decoder. We also introduce a *proxy loss* based on the silhouette, and show that the network can be trained to encode 3D shape using this 2D loss without explicitly including a 3D representation in the decoder. This is an advantage since the 2D loss is not then limited by the resolution of the 3D representation. To train the network we generate, and pre-train on, a large dataset of 'blobby' objects, and then fine tune *SilNet* on datasets for new tasks, such as for sculpture rendering.

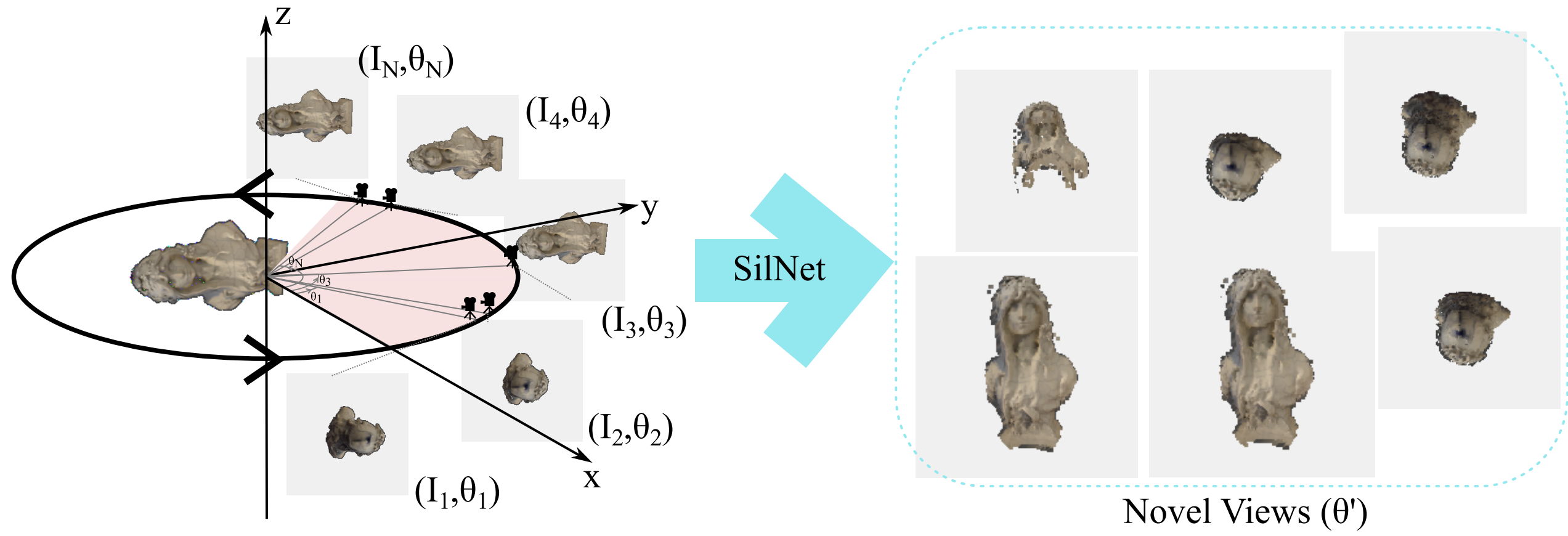

Figure 1: Each image and $\theta$ pair is processed jointly by SilNet to generate novel views of the sculpture. Zoom in for more detail.

Specifically, our contributions include the following. *First*, a new deep learning architecture (sec. 4) for multi-view shape encoding that handles a variable number of views at train and test-time in an order-agnostic manner; *Second*, two new synthetic datsts: a large scale dataset of simple blobby textured objects that can be used for pre-training the network, and a new challenging dataset of realistic sculptures with complex illumination and a wide variety of shapes (sec. 5). The sculpture dataset is used to demonstrate that our proxy task (sec. 3) is sufficient to learn and encode 3D shape, and enable silhouettes to be generated in new views for a wide variety of sculpture shapes and materials, as well as generating 3D representations; *Third*, we show (sec. 6) that using multiple views improves results over single views, and that the architecture is capable of generalising at test time to more or fewer views. The experiments also demonstrate the benefit of using the blobby object dataset for pre-training. Finally, we compare to the state of the art on the ShapeNet benchmark dataset [6].

## 2 Related work

There is a large body of work in the computer vision and graphics community on the area of reconstructing a 3D object, or generating new views of an object, given a set of images or silhouettes as input [5, 15, 22, 23, 25, 28, 31, 33, 34, 38]. Unlike modern approaches, these methods require multiple views of the object at test (inference) time to constrain the optimisation, and cannot predict parts of the object not visible in the input views (for example the back of the object if only the front is imaged). The exceptions are methods that employ strong prior information about the properties of a particular object class and consequently can proceed from a single view for inference [2, 3, 30].

The modern generation of deep learning approaches can be loosely divided into those that learn transformations on images to render new views, or those that generate 3D models. In both cases the learning has access to a large number of images of an object class (e.g. cars or chairs) that are usually synthetic. One of the first approaches for generating novel views based on one image using deep learning was that of Tatarchenko *et al.* [35] which could render chairs in novel viewpoints and interpolate between views. This approach was improved on to model the transformation of pixels as a flow field by Zhou *et al*. [41]. However, using an image based approach suffers from the problems of the smooth $L_1$ and $L_2$ losses which blur out the higher details. As a solution Park *et al*. [27] propose using GANs as the loss function.

A number of methods have been shown to be successful in generating simple 3D models from images of a set of classes. These approaches use priors based on the given object classes to fill in the unknown information using a deep network. For example the works by [10, 13, 19, 29, 32, 37, 39, 40] use deep learning architectures to learn a mapping from an image to voxels, point clouds, meshes, or geometry images. They differ in their choice of loss function (*e.g*. Euclidean, GANs, ray consistency, voxel-wise softmax etc.) and architecture design. The work of Moreno *et al*. [26] predicts a set of latent variables (*e.g*. the object class, lighting effects) using a neural network which is used by their generative model to render the scene.

Most of these works focus on generating objects from a single image. SilNet, on the other hand, handles a variable number of views at test time, and the results improve given more views. In this regard, SilNet is most similar to 3D-R2N2 [7] which learns a 2D to 3D mapping from images to voxels. 3D-R2N2 was also able to combine a variable number of views at test time, but it uses a RNN which is impacted by the order of the views and may forget salient information from the initial views. Rezende *et al*. [29] also considers generating models from multiple views, but they again use an LSTM which has the same limitations (and their architecture also receives additional information, as they use depth images as opposed to 2D rendered images). In contrast, our approach can input a variable number of views in an order-agnostic manner. Though the image-based approach of Zhou *et al*. [41] allows for the combination of multiple views, this is done following the decoder as each image generates its own prediction of the output image. They also train two separate models for the single/multi-view cases. In contrast, we explicitly allow the decoder access to the information from each view and SilNet handles a variable number of views at test time.

Moreover, many of the current approaches are class-specific [19, 26, 29, 32, 35] and require separate training for each class (car, chair, etc.). As a result, an important question is to what extent these architectures have actually learnt about shape in general. For example, Yan *et al*. [40] consider how their architecture performs on new categories unseen during training; they note that it performs as well on the new categories as the old only when the new categories are very similar to the original ones.

# 3 The silhouette: a proxy for learning to encode 3D shape

We consider the task of generating new views of a 3D object, given one or more images. In particular we ask the network to predict the object *silhouette* given only the angle of the new viewpoint. The key idea is that in order to carry out the task, the network will need to encode 3D shape, though we do not explicitly represent 3D shape or have a training loss on this. By concentrating only on the silhouette, the network does not need to learn to predict image intensities, so the learning (and inference) are easier. We also avoid the need for geometry images [32] or 3D voxel representations during training.

Geometrically, if the network can predict the silhouette for multiple views, then it must at least encode the *visual hull* [23] of the object – this is the maximum 3D object that is silhouette-equivalent to the given images. How well SilNet has encoded 3D shape can then be probed by asking it to create silhouettes of new objects at new viewpoints or by extracting the implicitly learned 3D shape. Our approach is inspired by the work of Koenderink on inferring 3D properties of smooth surfaces from their occluding contours [21].

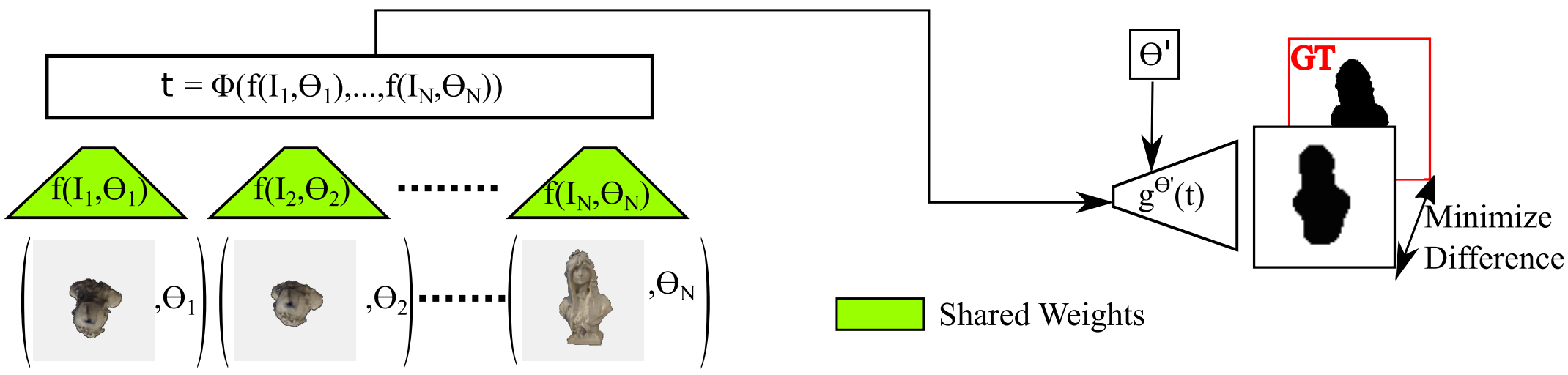

Figure 2: Training framework. Each image/$\theta$ pair is processed by a separate encoder $f$ to output a feature vector. These are combined (in our case, max-pooled) by $\phi$ to obtain a combined feature vector. This is decoded by $g$ to generate the silhouette of the object in a new view $\theta'$.

**Loss Function.** We are given a set of images $I_1 \dots I_N$ of an object taken at angles $\theta_1 \dots \theta_N$ (see fig 1) and ground truth silhouette $S$, where $S_{x,y} \in 0,1$ where 0 means object and 1 non-object. We wish to learn a function $g^{\theta'}$ that generates $S$ at an angle $\theta'$ (see fig 2). A pixel-wise binary cross-entropy loss $L$ minimises the difference between the ground truth and predicted silhouette: $L(S, g^{\theta'}) = \sum_{x,y} g^{\theta'}_{x,y} \log(S_{x,y}) + (1 - g^{\theta'}_{x,y}) \log(1 - S_{x,y})$.

# 4 The SilNet Architecture

In order to generate the silhouette from multiple views, SilNet uses the encoder/decoder architecture, outlined in figure 2. It consists of an *encoder* $f$, which can be replicated as many times as there are numbers of views. As the parameters of all the encoders are shared, this corresponds to no increase in memory. The encoders are combined in a *pooling layer* $\phi$ which max pools over the feature vectors for each tower to learn a combined feature vector. This allows SilNet to take into account multiple images, as it can attend to the important features from each image in the pooling layer. Finally a *decoder* $g^{\theta'}$ up-samples from the feature vector to generate a 2D image of the silhouette in a new viewpoint $\theta'$. A detailed overview of the architecture for one target is given in figure 3. The learned feature vector can also be used to generate a 3D latent shape representation using a 3D decoder. This latent shape is projected to a 2D image of the silhouette using a projection layer described below. Note, that due to the pooling layer the number of images used as input can vary (and indeed may differ between train/test time).

Each image and $\theta$ pair is encoded in a separate encoder tower. The images are resized to $112 \times 112$. The theta parameter is encoded as $(\sin\theta, \cos\theta)$ to represent a distribution of angles such that $0°$ is closer to $359°$ than $180°$. These theta values are passed through two fully connected layers, broadcast and concatenated to the corresponding tower.

In the decoder, the feature vector is up-sampled and followed by a pixel-wise sigmoid. An additional convolutional layer is added following the final two up-sampling layers [9].

## 4.1 3D Decoder

In order to extract the 3D object and ascertain whether the 2D features encode information about 3D shape, SilNet's decoder is modified such that SilNet learns a latent representation of the 3D shape while the encoder is kept fixed. In the 3D decoder, the combined feature vector is up-sampled using 3D convolutional transposes to generate a $57 \times 57 \times 57$ volume $V$ which is followed by a sigmoid layer (for full details, please see the extended paper). This volume can be imagined as a 3D representation of the object, which can be projected to obtain the silhouette using the projection layer described below. Finally, a binary cross entropy loss over the projected silhouette is used, as for the 2D decoder. Note, there is no direct loss on a 3D representation, as in many previous methods.

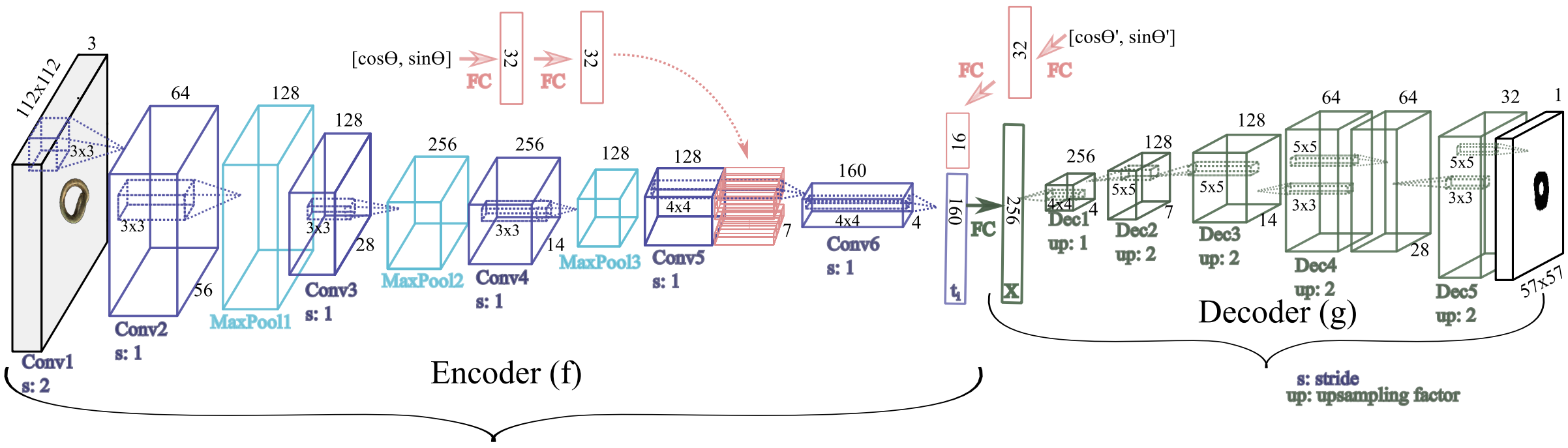

Figure 3: Silhouette prediction in the case of a single view. In the case of multiple input views, the feature vector $t_i$ resulting from the encoder for image $I_i$ is max-pooled over the multi-towers to give $t$, as shown in figure 2. This feature vector is up-sampled by the 2D decoder which is parameterised by $\theta'$. Convolutions and convolutional transposes are followed by ReLU units except for the last convolution which is followed by a pixel-wise sigmoid.

**Projection layer $T_{\theta'}$.** Given a voxel assumed to represent a 3D shape, we wish to project this to a 2D image to use our loss function over silhouettes. $V$ is first rotated by $\theta'$ using a nearest neighbour sampler [18]. Then, the min value over all depth values for each pixel location is used to determine whether the pixel is filled or not. Assuming orthographic projection and rotation $\theta'$ about $z$, the projected image pixel $p_{j,k}$ is given by

$$\begin{aligned} V_{\theta'}(i,j,k) &= V(\lfloor \cos(\theta')i - \sin(\theta')j + 0.5 \rfloor, \lfloor \sin(\theta')i + \cos(\theta')j + 0.5 \rfloor, k) \\ p_{j,k} &= \min_i V_{\theta'}(i,j,k) \end{aligned} \tag{1}$$

where $V_{\theta'}(i,j,k)$ denotes the rotated box. This is a differentiable composition of functions, so the pixel-wise classification loss can be back-propagated through this layer.

A similar layer was investigated by [12, 29, 37, 40]. Yan *et al.* [40] treat the silhouette as a regression problem and use a Euclidean loss. Tulisani *et al.* [37] use a ray potential to enforce consistency constraints. Gadelha *et al.* [12] treat the volume as opaque, using an exponential function to combine the summation of values in the volume at each pixel location followed by a GAN as their loss function. Rezende *et al.* [29] use a learned projection module which requires them to constrain the latent volume at train time using multiple output images. Our approach differs from previous work, as we use the the min function (as does [40]), and also we treat the silhouette as a binary classification problem (as opposed to a regression problem) which is simpler to train. We further demonstrate in our experiments (section 6) how we can achieve good results without latently generating a 3D shape, and we can incorporate additional views *at test time* to achieve superior results to the work of Yan *et al.* [40].

# 5 Datasets

Three synthetic datasets are used. Sample renderings from each are given in figure 4.

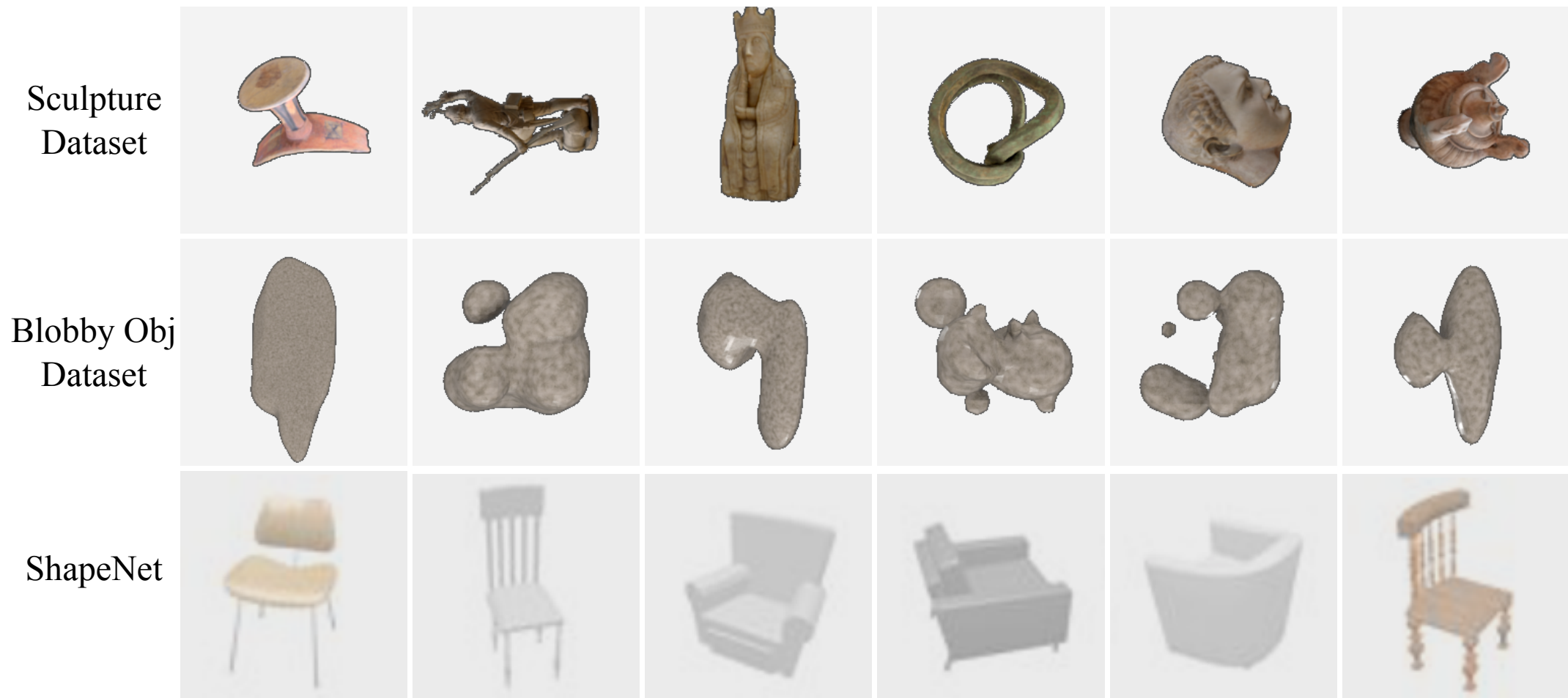

Figure 4: Samples from each dataset exhibiting the sculptures' variety and complexity.

**Blobby objects[1].** This consists of smooth (undulating) surfaces created using implicit surfaces. It contains 11,706 blobby objects split 75/10/15 into train/val/test. There are a very small number of images (five) per object so that SilNet must reason about 3D shape to solve the proxy task. The images are rendered under orthographic projection using Cycles in Blender [4]. Five views are created as follows: first, three light sources are randomly distributed about the object (and from view to view); second, the camera is rotated about the $z$-axis and the $\theta$ value for each rendering chosen uniformly at random from $[0^\circ, 120^\circ]$; finally, a complex texture model is used; it exhibits subsurface scattering and has diffuse/specular reflections. This dataset is also used for pre-training the network for other tasks.

**Sculpture dataset[1].** We compiled a new dataset of 307 realistic sculptures from SketchFAB [1]. Again, it is split 75/10/15 into train/val/test. Images are rendered for five views as done for the blobby objects. These sculptures have no canonical orientation unlike ShapeNet (*e.g.* in ShapeNet chairs are aligned such that $0^\circ$ corresponds to the chair facing the viewer).

**ShapeNet.** ShapeNet [6] is a large dataset of 3D models divided into sub-categories. To compare this work to Yan *et al.*'s [40], we use their subdivision, train/val/test split and renderings. Their renderings of ShapeNet objects are simpler than those of the sculpture dataset. Their objects have no complex reflectance properties, the lighting conditions are constant, and they render 24 views at fixed (not random) $15^\circ$ intervals about the $z$-axis for each object.

# 6 Experiments

This section evaluates the performance of SilNet on the blobby object dataset, the sculpture dataset, and also compares SilNet to the work of Yan *et al.* [40] on ShapeNet. We demonstrate that using multiple towers improves results in all of these scenarios.

**Evaluation measure.** Results are reported using the mean Intersection-over-Union (IoU) for the testing partition of the datasets. The IoU for a given predicted silhouette $S$ and ground truth silhouette $\bar{S}$ is defined as $\frac{\sum_{x,y}(I(S) \cap I(\bar{S}))}{\sum_{x,y}(I(S) \cup I(\bar{S}))}$ where $I$ is an indicator function that equals 1 if the pixel corresponds to an object. This is then averaged over all images for the mean IoU.

[1] Available at http://www.robots.ox.ac.uk/~vgg/data/SilNet/.

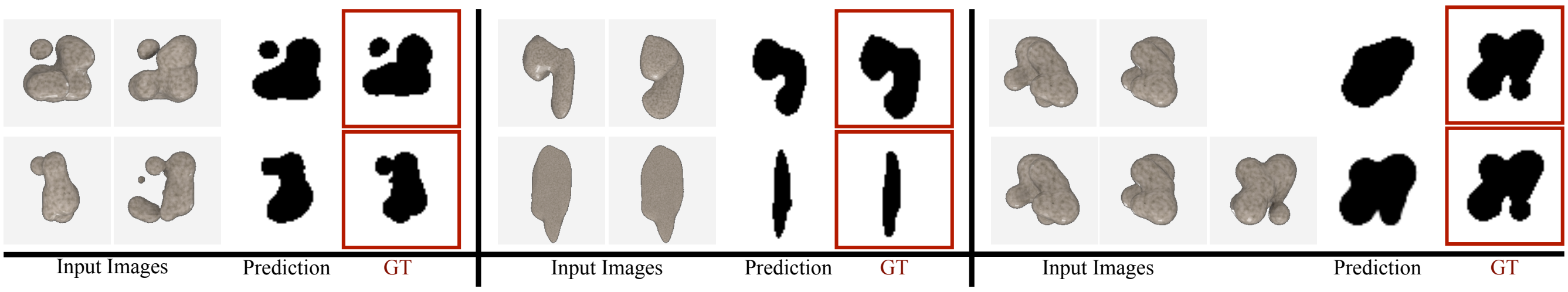

Figure 5: Sample silhouette predictions of SilNet trained with two towers on the blobby object dataset. The left hand images are the input views, the right-hand the ground-truth silhouette and the central ones SilNet's prediction. The rightmost column also demonstrates how increasing the number of input views improves the results.

**Training/Testing setup.** Datasets are divided randomly into the train/val/test splits such that objects are in one set. This ensures the generalisability of SilNet to unseen objects. When training with $N$ towers, $N+1$ views of an object are randomly selected. The mask of one of these views is used as the silhouette to be predicted, and the rest are given as input images. We also re-train SilNet with the entire train/val set when reporting results.

When comparing results for differing numbers of towers, we similarly randomly choose an object and $N+1$ views for $N$ towers. With each new tower we include an additional unselected view. We ensure these choices are consistent when comparing variants of SilNet.

The parameters of SilNet are initialised with Xavier initialisation [14]. For the blobby objects, SilNet is trained using SGD with a momentum of 0.9, weight decay of 0.001, and batch size of 16. The blobby object dataset is then used to initialise a network for fine-tuning on the sculpture dataset. The Adam solver [20] is used for training with a learning rate of $1e^{-5}$, $\beta_1 = 0.9, \beta_2 = 0.999$ for the 2D case and SGD with a momentum of 0.9, weight decay of 0.001, and a batch size of 32 for the 3D case. Data augmentation is included in the form of jittering the y-location of the object and subtracting off the mean image.

## 6.1 Blobby Objects

There are 5 views of an object in the train/test set at different, random viewpoints, so to generate the silhouette in new views, SilNet must implicitly understand the object's 3D shape. For example, a bump (*e.g.* a nose) may not impact the object boundary in the given view (*e.g.* the full face frontal view) but lies on the silhouette in the new view (*e.g.* the profile). To predict the silhouette, SilNet must recognise the bump and thereby understand the object's 3D shape.

We first consider how the architecture copes when the number of towers at training or testing time is varied. Results are given in table 1. It can be seen that the performance is high, and that SilNet trained with two or three towers, for example, does indeed generalise to one, two, or three towers at test time. The major reason SilNet performs better given more views at test time is the problem of self-occlusion (fig 5). Continuing the example above, from the back of the head the nose is hidden, but it is visible in the silhouette corresponding to the profile view. With more views, it is more likely that the hidden part of the object (the nose) will be partially visible, allowing SilNet to construct the silhouette in the new viewpoint.

We found that max pooling outperforms average pooling when combining towers (see table 1). Max-pooling allows the combined feature vector to jointly record information about the viewpoint and associated embedding enabling a more direct path from the output viewpoint to the relevant (*e.g.* closer) input image embeddings. In the average pooling case, this viewpoint information is averaged out and lost. Using minor modifications to the approach described by Mahendran and Vedaldi [24], we can visualise the different properties of the two networks (fig 6). Given the inputs, we run a forward pass through the network and extract vector $X$ (the feature vector following the FC layer and adding the output angle $\theta'$ – see fig 3). We then fix the angles and generate the most likely input images starting from noise that will regenerate $X$. Given the three ground truth viewpoints, SilNet with max pooling can reconstruct the original input images, as the association between image embeddings and angles is kept. However, SilNet with average pooling struggles to reconstruct the input images.

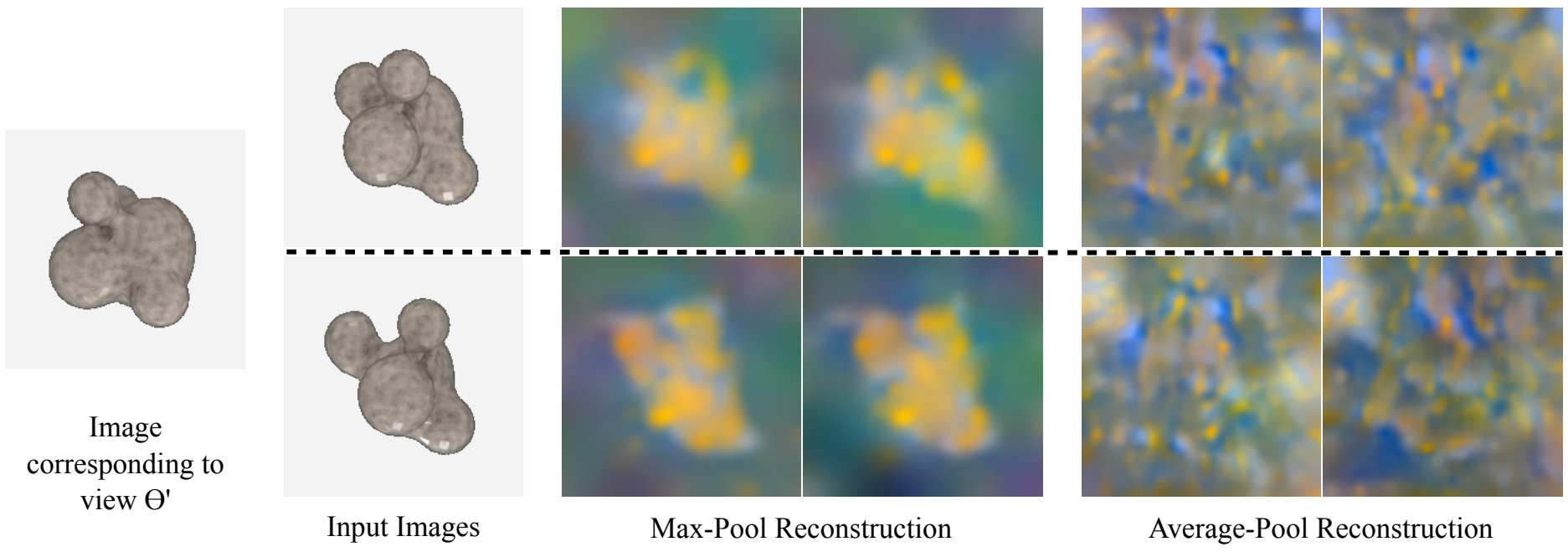

Figure 6: Reconstructions of the original input images when generating the new viewpoint shown (left), using a method similar to [24]. The two reconstructions per viewpoint demonstrate different random initialisations. It is clear that the network is more easily able to reconstruct the original images using the max pooling architecture, implying that it records a joint embedding of image and viewpoint when generating the new image.

| | Num. of towers tested with | | |
|---|---|---|---|
| | 1 | 2 | 3 |
| trained with 1 tower | 0.833 | 0.832 | 0.830 |
| trained with 2 towers | 0.885 | 0.927 | **0.934** |
| trained with 3 towers | 0.872 | 0.925 | **0.936** |
| avg (2 towers) | 0.883 | 0.895 | 0.894 |

Table 1: Mean IoU of variants of SilNet on the blobby object dataset. Higher is better. The top three variants use max (not average) pooling to combine feature vectors.

| *Id* | # towers trained with | pre-training | Num. of towers tested with | | | |
|---|---|---|---|---|---|---|
| | | | 1 | 2 | 3 | 4 |
| **SilNet$^{2D}$** | | | | | | |
| (A) | 1 View | blobby objects | **0.755** | 0.780 | 0.780 | 0.775 |
| (B) | 2 Views | – | 0.703 | 0.746 | 0.758 | 0.761 |
| (C) | 2 Views | blobby objects | 0.735 | **0.821** | **0.832** | **0.834** |
| (D) | 3 Views | blobby objects | 0.720 | 0.815 | **0.830** | **0.836** |
| **SilNet$^{3D}$** | | | | | | |
| (E) | 2 Views | – | 0.711 | 0.745 | 0.751 | 0.755 |
| (F) | 2 Views | † | **0.732** | 0.770 | 0.773 | 0.776 |
| (G) | 2 Views | blobby objects | 0.713 | **0.777** | **0.788** | **0.793** |

Table 2: Mean IoU of variants of SilNet on the sculpture dataset. Higher is better. † indicates the encoder from SilNet$^{2D}$ (C) was used and frozen during training.

## 6.2 Sculpture Dataset

**Experiments.** Results are given in table 2. We first demonstrate that a curriculum learning strategy of pre-training on the blobby object dataset and then fine-tuning on the sculpture dataset improves results in both the 2D case (SilNet$^{2D}$) and 3D case (SilNet$^{3D}$). Compare row B to row C for example in the 2D case. Here the 3D case refers to the architecture which uses 3D convolutions and the projection module from which we can recover a latent 3D shape, and the 2D case refers to when the shape is only learnt implicitly – sec. 4. Second, we demonstrate that the features learned in the 2D case for the sculpture dataset generalise to the 3D case. More precisely, we fix the encoder trained in the 2D case, and train the 3D decoder separately. The results of this (row F) are better than training SilNet$^{3D}$ from scratch (row E). This implies that the feature vectors learned in the 2D case encode something about 3D shape, and also that the pre-training for F is better than the no pre-training of E. However, we note that the 2D case consistently outperforms the 3D case. We hypothesise this is due to the 2D decoder having fewer weights to learn – simplifying training – and not being forced to explicitly represent the 3D shape – allowing for flexibility. Finally, the results demonstrate quantitatively that SilNet trained using $N$ towers generalises to more/fewer towers at test-time; this is demonstrated visually in figure 7. The improvement with more views is again a result of self occlusion, as in the blobby object dataset.

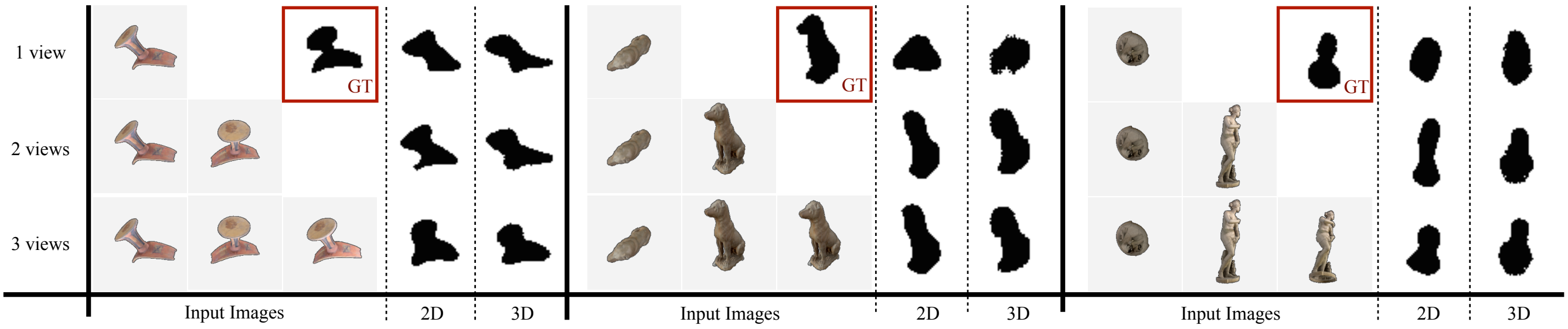

Figure 7: Silhouette predictions for SilNet$^{2D}$(C)/SilNet$^{3D}$(G) trained on two views on the sculpture dataset. The left hand images are the input test views, the ones boxed in red the ground-truth silhouettes and the right-hand ones SilNet$^{2D}$/$^{3D}$'s predictions. Each row corresponds to adding an additional view, which improves SilNet's predictions.

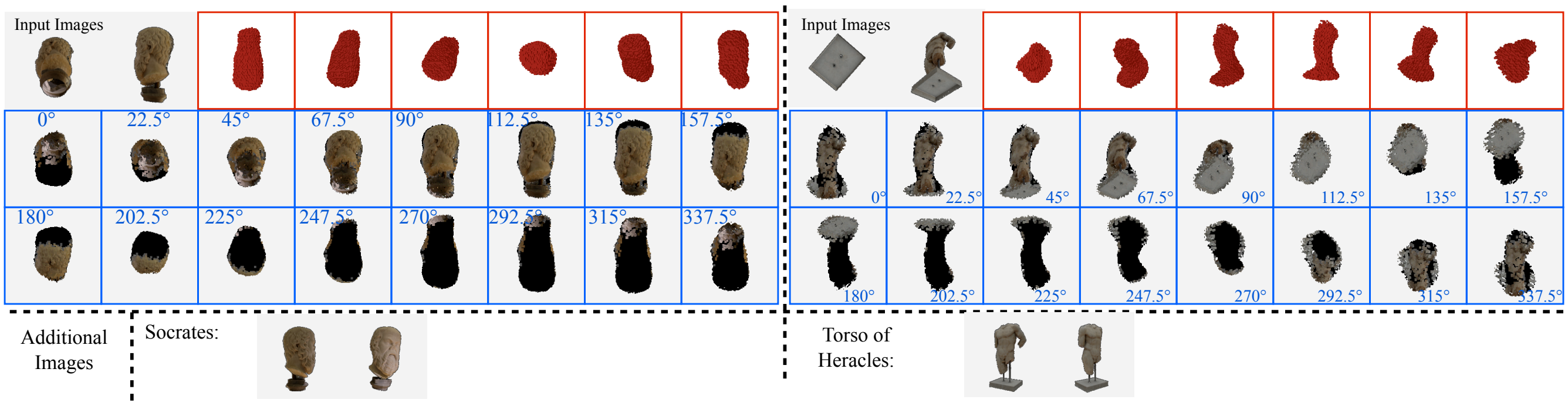

Figure 8: SilNet$^{3D}$'s predictions for the given input images. Additional images in more natural viewpoints are provided to orient the reader. The latent 3D volumes $V$ are shown in red. These are up-sampled 3× and rendered using view dependent texturing [8]. The renderings are given in the blue squares (the black pixels indicate that this portion of the sculpture is not visible in the given images). SilNet is trained on images with $\theta/\theta' \in [0°, 120°]$, so it has to extrapolate for $\theta' > 120°$. Zoom in for details.

**An application: novel view synthesis.** Although generating 3D models is not the thrust of this work, we exhibit how well SilNet$^{3D}$(G) extrapolates to novel views in figure 8. Please see the extended paper for more examples.

## 6.3 ShapeNet

In this section, the generalisability of the multi-tower portion of SilNet is compared against the architecture of Yan *et al.* [40] on the ShapeNet chair test set and we demonstrate SilNet performs better. We only compare to Yan *et al.* [40] as theirs is the only publicly available state-of-the-art model that uses segmentation masks to infer 3D shape from one image without strong priors on the object class. Their silhouette $\bar{S}$ is a real-valued mask corresponding to how likely a pixel is to be object or background. They first train their encoder to differentiate between different classes and then train the decoder for the given class.

To compare, SilNet$^{2D}$ with two towers is trained on their full set of train classes. SilNet was trained from scratch on the entire ShapeNet train subset. One variant of SilNet was then fine-tuned on the chair subset, which is the same manner that Yan *et al.* [40] use to train their model. To use SilNet's classification loss, the publicly available masks from [40] are thresholded at $\bar{S}_{x,y} > 0.5$. To compare the generated silhouettes, we use their implementation of the IoU metric $\frac{\sum_{x,y} I(S_{x,y}) \times \bar{S}_{x,y}}{\sum_{x,y}((I(S_{x,y})+\bar{S}_{x,y})>0.9)}$. Table 3 compares the results and demonstrates that SilNet is comparable given only one view (though it was trained with two) but performs better with more views. Some visual examples are given in figure 9. Interestingly, given more views, SilNet without fine-tuning performs better than SilNet fine-tuned on the chair set, implying that more data and more views yield better generalisability.

| | Fine-tuned on: | Num. of towers tested with | | | | |
|---|---|---|---|---|---|---|
| | | 1 | 2 | 3 | 4 | 5 |
| **Yan *et al.* [40]** | chair | 0.797 | – | – | – | – |
| **SilNet$^{2D}$** | chair | 0.792 | 0.806 | 0.809 | 0.809 | 0.809 |
| **SilNet$^{2D}$** | – | 0.788 | 0.818 | 0.828 | 0.833 | 0.835 |

Table 3: Mean IoU on the chair ShapeNet class. Higher is better. SilNet is comparable in the 1-view case but performs consistently better given more views.

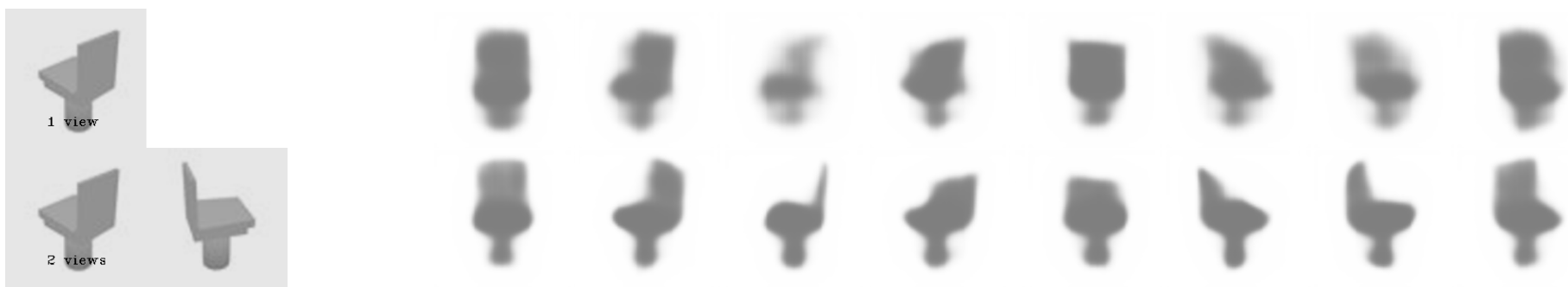

Figure 9: Performance of SilNet$^{2D}$ on the ShapeNet chair test set. The input images (and angles) are kept constant while the angle corresponding to the output view $\theta'$ is rotated between $[0°, 360°]$. The input images are the coloured images to the left and each row corresponds to the addition of another view, which improves SilNet's predictions.

## 7 Conclusion

We have demonstrated a novel architecture, SilNet, for performing 3D shape understanding using neural networks that works in challenging scenarios (*e.g.* a small number of images, wide baselines, and complex illumination). The proxy task of producing silhouettes in new views forces SilNet to encode 3D shape without ever having to see a ground-truth 3D volume or even an object in more than five views from a restricted azimuth range. Moreover, SilNet trained with $N > 1$ towers is able to combine the information from multiple images at test time and performance improves accordingly. Both SilNet$^{2D}$/$^{3D}$ generate visually compelling and consistent silhouettes in new views on our challenging and realistic sculpture dataset. Interestingly, SilNet$^{2D}$ outperforms SilNet$^{3D}$ on our proxy task implying improvements to the naive 3D decoder should be investigated (*e.g.* [16, 36]).

**Acknowledgements.** This work was funded by an EPSRC studentship and EPSRC Programme Grant Seebibyte EP/M013774/1.